\documentclass{article}
\usepackage{spconf,amsmath,graphicx}
\usepackage{amssymb}
\usepackage{multirow}
\usepackage{subcaption}
\usepackage{hyperref}
\usepackage{comment}

\title{Quantization Effects on Neural Networks Perception: How would quantization change the perceptual field of vision models?}
\name{Author(s) Name(s) \thanks{Thanks to TIC-ART Regional Project for funding this research.}}
\address{Author Affiliation(s)}
\name{Mohamed Amine Kerkouri $^{1*}$\thanks{* The first two authors have equal contributions.}, Marouane Tliba$^{1*}$, Aladine Chetouani$^{1}$ and Alessandro Bruno$^{2}$}

\address{$^{1}$Laboratoire PRISME, Université d'Orléans, Orléans, France\\
$^{2}$IULM University , Milan, Italy
\\}
\begin{document}
%\ninept
%

% \IEEEoverridecommandlockouts

\maketitle
%
% Title: Quantization Effects on Perception: A Comprehensive Study across Diverse Neural Network Architectures
% Abstract:

\begin{abstract}  
Neural network quantization is a critical technique for deploying models on resource-limited devices. Despite its widespread use, the impact of quantization on model perceptual fields, particularly in relation to class activation maps (CAMs), remains underexplored. This study investigates how quantization influences the spatial recognition abilities of vision models by examining the alignment between CAMs and visual salient objects maps across various architectures. Utilizing a dataset of 10,000 images from ImageNet, we conduct a comprehensive evaluation of six diverse CNN architectures: VGG16, ResNet50, EfficientNet, MobileNet, SqueezeNet, and DenseNet. Through the systematic application of quantization techniques, we identify subtle changes in CAMs and their alignment with Salient object maps. Our results demonstrate the differing sensitivities of these architectures to quantization and highlight its implications for model performance and interpretability in real-world applications. This work primarily contributes to a deeper understanding of neural network quantization, offering insights essential for deploying efficient and interpretable models in practical settings.

\end{abstract}


\vspace{-3mm}

\section{Introduction}
\label{sec:intro}
\vspace{-3mm}

The rapid advancements in deep learning have led to the development of increasingly complex neural network architectures. These models have demonstrated remarkable performance across various tasks, such as image classification, object detection, and segmentation. However, deploying these models on resource-constrained devices remains challenging due to their high computational and memory requirements.

Neural network quantization \cite{SurveyQuant} has emerged as a key technique for addressing the deployment challenges by reducing model size and computational complexity. 

Quantization involves converting the weights and activations of a deep neural network from floating-point precision to lower bit-width representations. Quantization techniques have evolved from the following equation:
\begin{equation}
    Q(x) = \text{round}\left(\frac{x}{\Delta} \right) \times \Delta
    \vspace{-2mm}
\end{equation}
where $x$ is the continuous value, $Q(x)$ represents the quantized value, and $\Delta$ is the quantization step size, determining the precision of the quantized representation.\\

By reducing the precision of parameters, quantization enables the delivery of lighter-weight models for efficient execution on devices with limited computational resources, such as smartphones \cite{quant_mobile}, embedded systems \cite{quant_embed}, and IoT devices \cite{quant_iot}.

While quantization offers practical benefits, it introduces changes to model behavior that may impact interpretability. Interpretable models are crucial for understanding the decision-making process of neural networks, particularly in safety-critical applications such as medical diagnosis and autonomous driving. Ensuring that quantized models maintain interpretability is essential for building trust and facilitating human-machine collaboration.

Class activation maps (CAMs) \cite{cams} are visualization techniques that highlight the regions of an image contributing most to a neural network's output. CAMs provide valuable insights into where the model focuses its attention when making predictions, enhancing understanding and interoperability. Some algorithms like Grad-CAM \cite{gradcam} and Grad-CAM++ \cite{gradcampp} analyze the gradients flowing back into the final convolutional layer of a CNN. These gradients highlight which features in the image contributed most to the model's decision, and the algorithm uses them to create a CAM pinpointing these important regions. Understanding how quantization affects CAMs can provide insights into the changes in model behavior and decision-making processes.

Visual saliency \cite{Koch1987} refers to the perceptual prominence or importance of different regions within an image for the human visual system. Unlike CAMs, which are derived from the internal activations of a neural network, visual Saliency maps and salient  objects maps is based on human perception and attention. Comparing CAMs with visual salient object maps \cite{zheng2024birefnet} can help assess the alignment between model attention and human cognitive perception to evaluate the interpretability of quantized models.

In this paper, we investigate the impact of quantization on model perception, focusing on CAMs and their alignment with visual salient objects maps. We evaluate six diverse neural network architectures on a comprehensive dataset of images from ImageNet \cite{imagenet_cvpr09}. Through rigorous experimentation, we analyze the changes in CAMs induced by quantization and their implications for model interpretability and performance.

Our findings contribute to advancing the understanding of neural network quantization and offer insights crucial for deploying efficient and interpretable models in practical settings.\footnote{\href{https://github.com/kmamine/CAM_Quant}{Source Code}}

\vspace{-5mm}
\section{Methodology}
\label{sec:method}
\vspace{-3mm}

\begin{figure}
    \centering
    \includegraphics[width=0.8\linewidth]{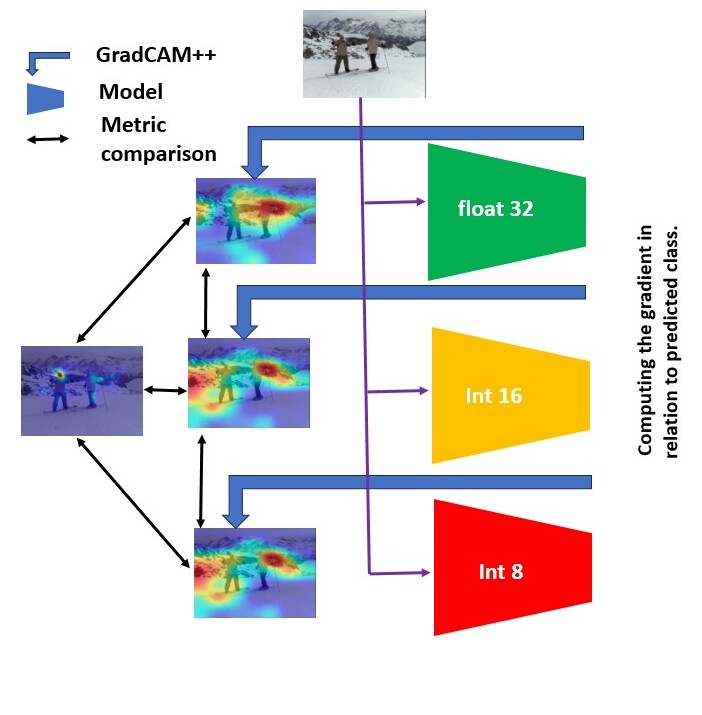}
    \caption{The experimental methodology of experimentation.}
    \label{fig:enter-label}
    \vspace{-6mm}
\end{figure}

In this study, we aim to explore the effects of quantization on neural network model perception. By investigating how quantization affects CAMs across different architectures, we hope to uncover insights that can enhance model efficiency and interpretability in real-world applications.

\subsection{Model Selection and Quantization}

We selected six widely-used neural network architectures, namely VGG16 \cite{vgg}, ResNet50 \cite{resnet}, EfficientNet \cite{efficientnet}, MobileNet \cite{mobilenetv2}, SqueezeNet \cite{squeezenet}, and DenseNet \cite{densenet}, due to their diverse structures and widespread adoption across various computer vision tasks.

Quantization, the focal subject of our study, was performed using the Quantization Aware Training (QAT) module from PyTorch\footnote{\href{https://pytorch.org/docs/stable/quantization.html}{Quantization PyTorch}} only for inference, in order to simulate adaptive quantization. This technique allows for the quantization of neural network weights and activations while preserving the gradient flow necessary for calculating CAMs. Three quantization levels were chosen: Float 32 (\textbf{f32}), Integer 16 (\textbf{int16}), and Integer 8 (\textbf{int8}). It is worth noting that other quantization techniques in PyTorch do not support gradient flow during quantization. 

In PyTorch, adaptive quantization is often implemented using Quantization-Aware Training (QAT), which involves simulating quantization effects during training while keeping the gradient flow intact as mentioned before. 
% Here, I'll provide a more specific equation that captures the adaptive nature of quantization typically employed in frameworks like PyTorch.

The adaptive quantization process in PyTorch can be described using the following equations:

\begin{enumerate}
    \item \textbf{Scale and Zero Point Calculation:}  For adaptive quantization, the scale \(s\), which  is a factor used to convert a floating-point value into an integer value, and zero point \(z\), which is an integer value that maps the minimum floating-point value \(x_{min}\)  to the minimum quantized value \(q_{min}\),   are calculated based on the minimum and maximum values of the input tensor:

   \begin{equation}
       s = \frac{x_{\text{max}} - x_{\text{min}}}{q_{\text{max}} - q_{\text{min}}}
   \end{equation}
   
   \begin{equation}
   z = q_{\text{min}} - \frac{x_{\text{min}}}{s}
   \end{equation}

where \(x_{\text{min}}\) and \(x_{\text{max}}\) are the minimum and maximum values of the input tensor \(x\).
\(q_{\text{min}}\) and \(q_{\text{max}}\) are the minimum and maximum values of the quantized range (e.g., for int8, \(q_{\text{min}} = 0\) and \(q_{\text{max}} = 255\)).

      \item \textbf{Quantization Equation:} 
The quantization of a value \(x\) is performed as follows:

\begin{equation}
   Q(x) = \text{round}\left(\frac{x}{s} + z\right)
\end{equation}

\end{enumerate}

Combining these steps, the full adaptive quantization process can be summarized:

\begin{equation}
Q(x) = \text{round}\left(\frac{x - x_{\text{min}}}{\frac{x_{\text{max}} - x_{\text{min}}}{q_{\text{max}} - q_{\text{min}}}} + q_{\text{min}} \right)
\end{equation}

% In PyTorch's Quantization-Aware Training (QAT), this process is integrated into the model training loop, ensuring that the quantization effects are taken into account while updating the model weights. This approach helps in preserving the model accuracy even after the quantization.

\subsection{CAM Calculation}

CAMs were computed for each image in the subset chosen from the validation set of ImageNet \cite{imagenet_cvpr09} across different quantization levels. CAMs provide insights into the regions of an image that contribute most to a neural network's decision-making process, facilitating interpretability and understanding of model behavior. The Grad-CAM++ \cite{gradcampp} method was employed for CAM calculation, leveraging its ability to capture gradient information effectively and produce accurate heatmaps.

\subsection{Visual Salient Object Maps}
% \vspace{-3mm}

Visual Salient Object Maps were generated using the BiRefNet  model \cite{zheng2024birefnet}, a state-of-the-art technique for detecting visual salient objects that showed promising results for both high and low-level scenes. Visual saliency refers to the perceptual prominence or importance of different regions within an image according to human perception and attention. While visual salient objects are the set of objects that represent the most prominent items depicted in the scene, aiming to segment the foreground from the background of the scene. 
By comparing CAMs with visual salient objects maps, we aim to evaluate the alignment between model attention and human perception across different quantization levels.

%\subsection{Evaluation Metrics}
% \vspace{-3mm}
\subsection{Evaluation Metrics}

To assess the similarity or difference between different CAMs at different quantization levels and between CAMs and visual salient objects maps, we employed three metrics \cite{sal_metrics}: Similarity (SIM), Kullback-Leibler Divergence (KLD), and Pearson Correlation (CC).

\textbf{Similarity (SIM)}: Known as histogram intersection, this metric is used to measure the similarity between two distributions. It was introduced in \cite{SIM} for the task of image matching.

\begin{equation}
\begin{split}
        SIM(S_{GT},S_{CAM}) = \sum_{i} \min(S_{GTi}, S_{CAMi}) \hspace{5mm}
        \\ \mbox{ where }  \hspace{5mm}
        \sum_{i}S_{GTi} = \sum_{i}S_{CAMi} =1
        \label{eq:SIM}
\end{split}
\end{equation}
Where $S_{GT}$ represnets the salient object mask, and $S_{CAM}$ represents the Class Activation Map. \\

\textbf{Kullback-Leibler Divergence (KLD)}: This is a general information-theory-based metric that measures the difference between two probability distributions.

    \begin{equation} 
        KL(S_{GT},S_{CAM}) = \sum_{i}S_{CAMi} \log\left(\epsilon + \frac{S_{CAMi}}{\epsilon+S_{GTi}}\right)
        \label{eq:KL}
    \end{equation}

\textbf{Pearson Correlation (CC)}: Also known as the linear (Pearson) correlation coefficient, this metric is used for measuring the dependency between two heatmaps. A high correlation between the maps indicates that the two distributions are closely related and interconnected.

    \begin{equation} 
        CC(S_{GT},S_{CAM}) =  \frac{Cov(S_{GT},S_{CAM})}{\sigma(S_{GT})\times\sigma(S_{CAM})}
    \label{eq:CC}
    \end{equation}

We conducted experiments on a dataset consisting of 10,000 images randomly selected from the validation set of ImageNet \cite{imagenet_cvpr09}. Each neural network architecture was quantized at the specified levels using QAT, and CAMs were computed for each image in the dataset. %Visual Saliency Maps were generated using the SATSal model for comparison with CAMs.

This comprehensive methodology enabled us to systematically investigate the impact of quantization on model perception and interpretability, providing insights into the behavior of neural networks across different quantization levels and their alignment with human perception.

\begin{table*}
    \centering
    \begin{tabular}{cccccccc}
    \hline
        \textbf{Model}& \textbf{Metric}  & \textbf{VGG-16}  & \textbf{ResNet-50} & \textbf{DenseNet-121} & \textbf{MobileNet-V2} &  \textbf{SqueezeNet-1.0} & \textbf{EfficientNet-B0}   \\
        \hline
        \multirow{3}{*}{\textbf{f32 v. GT}} 
         & SIM $\uparrow$ & $0.44 \pm 0.19$ & $0.35 \pm 0.17$ & $0.37 \pm 0.21$ & $0.40 \pm 0.22$ &  $0.43 \pm 0.19$ &  $0.41 \pm 0.20$ \\
         & CC $\uparrow$ &  \underline{$0.43\pm 0.27$} &  \underline{$ 0.22 \pm 0.19$} &  $0.31 \pm 0.22$ &  $0.40 \pm 0.20$ &  \underline{$0.40 \pm 0.28$} &  \underline{$0.37 \pm 0.25$} \\
         & KLD $\downarrow$ &  \underline{$1.17 \pm 0.89$} &  \underline{$1.53 \pm 0.97$} &  \underline{$1.35 \pm 0.96$} &   \underline{$1.24 \pm 0.95$} &   $1.17 \pm 0.89$ &  \underline{$1.29 \pm 0.97$} \\
        \hline
        \multirow{3}{*}{\textbf{int16 v. GT}} 
         & SIM $\uparrow$ &  \underline{$0.67 \pm 0.33$} &  \underline{$0.64 \pm 0.36$}  &  \underline{$0.67 \pm 0.35$} &  \underline{$0.68 \pm 0.34$} & $0.43 \pm 0.19$ & \underline{$0.67 \pm 0.33$}\\
         & CC $\uparrow$ & $0.38 \pm 0.27$ & $0.21 \pm 0.19$ & $0.31 \pm 0.22$ & $0.40 \pm 0.20$ & $0.40 \pm 0.29$ & $0.36 \pm 0.25$ \\
         & KLD $\downarrow$ & $1.29 \pm 0.97$ & $1.63 \pm 1.03$ & $1.36 \pm 0.94$ &  $1.22 \pm 0.91$  & $1.17 \pm 0.89$ & $1.32 \pm 1.01$ \\
         \hline
        \multirow{3}{*}{\textbf{int8 v. GT}} 
         & SIM $\uparrow$ & $0.66 \pm 0.33$ & $0.57 \pm 0.36$ & $0.56 \pm 0.34$ &  $0.64 \pm 0.34$ & $0.43 \pm 0.19$ & $0.63 \pm 0.34$ \\
         & CC $\uparrow$ & $0.38 \pm 0.28$ & $0.20 \pm 0.19$ & $0.29 \pm 0.22$ &  $0.40 \pm 0.20$ & $0.40 \pm 0.29$ & $0.30 \pm 0.25$ \\
         & KLD $\downarrow$ & $1.29 \pm 0.97$ & $1.65 \pm 1.05$  & $1.37 \pm 0.95$ &  $1.24 \pm 0.92$ & $1.17 \pm 0.90$ & $1.40 \pm 1.00$ \\
         \hline\\
         \hline
        \multirow{3}{*}{\textbf{int16 v. f32}} 
         & SIM $\uparrow$ & \underline{$0.93 \pm 0.09$} & \underline{$0.96 \pm 0.07$}  & \underline{$0.96 \pm 0.07$}  & \underline{ $0.94 \pm 0.06$}  & \underline{$0.93 \pm 0.09$} &  \underline{$0.88 \pm 0.11$} \\
         & CC $\uparrow$ & $0.88 \pm 0.16$  & \underline{$0.94 \pm 0.17$}  & \underline{$0.90 \pm 0.22$}  &  \underline{$0.83 \pm 0.10$}  & \underline{$0.92 \pm 0.18$} & \underline{$0.75 \pm 0.15$} \\
         & KLD $\downarrow$ & $0.08 \pm 0.10$ & \underline{$0.05 \pm 0.13$}  & \underline{$0.05 \pm 0.11$}  &  \underline{$0.05 \pm 0.03$}  & \underline{$0.05 \pm 0.11$} & \underline{$0.18 \pm 0.10$} \\
         \hline 
        \multirow{3}{*}{\textbf{int8 v. f32}} 
         & SIM $\uparrow$ & $0.92 \pm 0.09$ & $0.84 \pm 0.15$ & $0.86 \pm 0.11$ & $0.92 \pm 0.07$ &  $0.90 \pm 0.09$ & $0.81 \pm 0.17$\\
         & CC $\uparrow$ &  $0.88 \pm 0.16$ & $0.61 \pm 0.29$ & $0.56 \pm 0.26$ &  $0.76 \pm 0.12$ & $0.89 \pm 0.20$ & $0.46 \pm 0.24$\\
         & KLD $\downarrow$ & $0.08 \pm 0.10$ & $0.29 \pm 0.22$ & $0.18 \pm 0.12$ &  $0.08 \pm 0.04$ & $0.06 \pm 0.11$ & $0.40 \pm 0.17$\\
         \hline
    \end{tabular}
    \caption{Quantitative Results for models: \textbf{GT} represents the salient object map, \textbf{f32} represents the float 32 precision model, \textbf{int16} represents the integer 16 precision model, \textbf{int8} represents the integer 8 precision model}
    \label{tab:quant_results}
    \vspace{-3mm}
\end{table*}

\begin{figure*}
    \centering
    \includegraphics[width=1\linewidth]{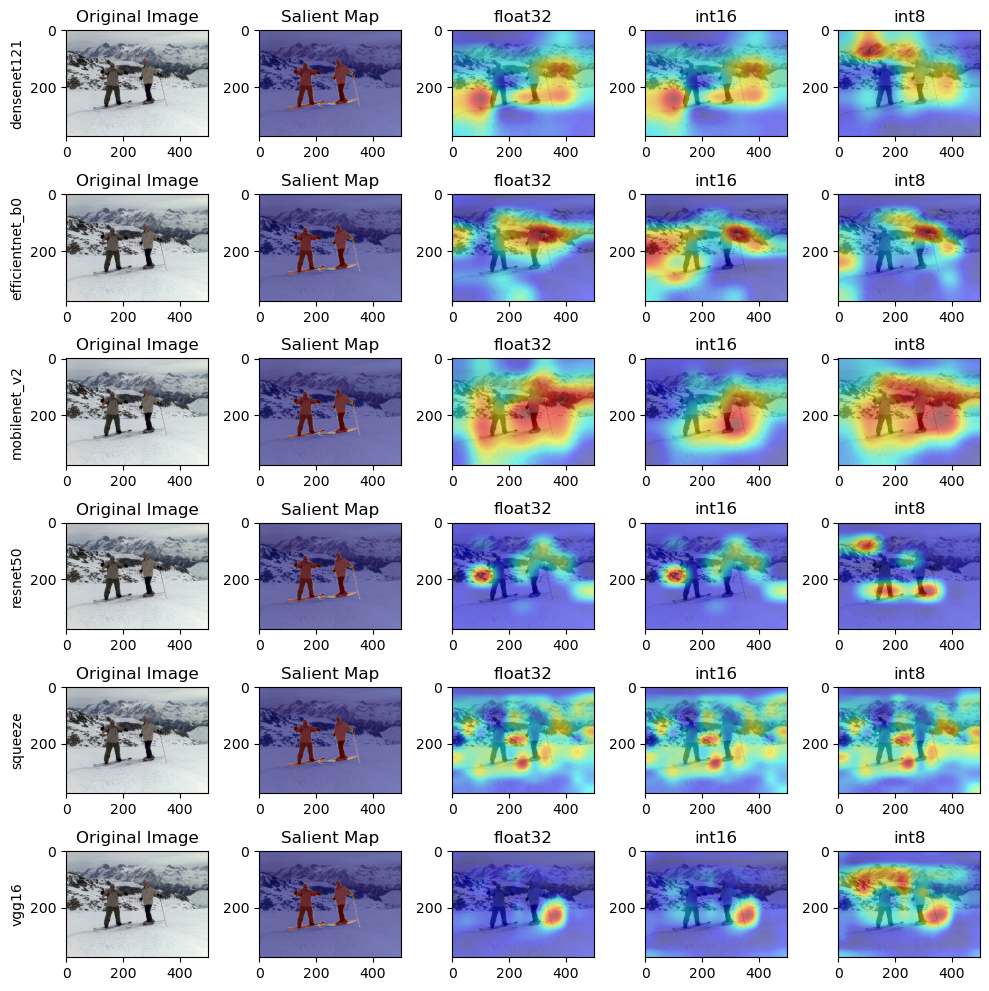}
    \caption{Qualitative results of CAMs and visual Salient object across multiple precisions and models.  }
    \label{fig:Qualt}
\end{figure*}

\section{Results}
\label{sec:results}
\vspace{-3mm}

The results highlight significant alterations in Class Activation Maps (CAMs) across diverse popular architectures under quantization. Our analysis unveils subtle shifts in model perception, revealing varying sensitivities among different architectures to quantization effects. A summary of our findings is presented in Table \ref{tab:quant_results}.

\subsection{Empirical Results}

The table \ref{tab:quant_results} summarizes the performance metrics across various neural network architectures (\textbf{VGG-16}, \textbf{ResNet-50}, \textbf{DenseNet-121}, \textbf{MobileNet-V2}, \textbf{SqueezeNet-1.0}, \textbf{EfficientNet-B0}) under different quantization levels (\textbf{f32}, \textbf{int16}, \textbf{int8}). It evaluates the similarity (SIM), Pearson correlation coefficient (CC), and Kullback-Leibler divergence (KLD) metrics for each combination.

% \textbf{Analysis}

\textbf{Similarity (SIM)}: Across all architectures, quantization generally increases SIM compared to ground truth (GT), especially noticeable for \textbf{int16} and \textbf{int8} levels. \textbf{VGG-16}, \textbf{ResNet-50}, \textbf{DenseNet-121}, and \textbf{MobileNet-V2} consistently show substantially higher SIM values compared to other models under the same quantization levels.  
The increase is due to the nature of the metric, as quantization may adjust the distributions of the salient object map and such that the minima operation in SIM captures more coinciding foreground regions between the two maps.

\textbf{SqueezeNet-1.0} shows high robustness in results. This can be attributed to the complex Fire modules that employs a squeeze and expansion operations to reduce and refine the amount of data. The model also bolsters complex skip connections that induce the propagation of information from shallow to deeper layers, decreasing loss of data, thus  introducing an inherent robustness to information loss induced by methods for adaptive quantization. 

\textbf{Pearson Correlation (CC)}: CC values vary significantly across architectures and quantization levels.
Higher correlations (\textgreater 0.5) are observed mainly between \textbf{f32} and \textbf{int16} levels, as evident in \textbf{ResNet-50}, \textbf{DenseNet-121}, and \textbf{SqueezeNet-1.0}. Lower correlation  is observed in the quantization of \textbf{EfficientNet-B0}. This is attributed to the more simplistic architectural design relying on straightforward sequence of convolutional layers while minimizing the number of parameters ($5.3M$ parameters). This leads increased sensitivity compared to other model that employ more robust methods to propagate information from shallow to deep layers using skip connections (e.i. \textbf{ResNet-50}, \textbf{DenseNet-121}) or other models that use much larger number of parameters like \textbf{VGG-16} ($138M$ parameters). The general trend indicates lower correlations  at higher quantization levels, indicating greater divergence from ground truth maps.

\textbf{Kullback-Leibler Divergence (KLD)}: KLD values generally decrease with higher quantization precision (\textbf{f32} to \textbf{int8}), suggesting improved alignment with ground truth Salient object maps.
\textbf{ResNet-50}, \textbf{DenseNet-121}, and \textbf{EfficientNet-B0} exhibit notable difference in KLD with increased quantization precision. These architectures are designed to be deeper making them more sensitive to  information alteration induced by quantization during forward propagation especially at lower precision levels (i.e \textbf{Int8}) . 

% Comparison Between Quantization Levels:

Comparing \textbf{int16} and \textbf{int8} with \textbf{f32} shows varying degrees of similarity and correlation changes across architectures.
Generally, \textbf{int16} maintains higher SIM and CC compared to \textbf{int8}, with exceptions observed in \textbf{SqueezeNet-1.0}. 

% \textbf{Interpretation}

Results indicate that different neural network architectures react differently to quantization, with some models maintaining higher similarity and correlation with ground truth even under lower precision.

Higher SIM and CC values suggest that certain architectures preserve visual interpretation capabilities better across different quantization levels, crucial for applications requiring accurate and reliable model interpretation.

This indicates that Therefore, while the number of parameters can be indicative of model complexity and potentially its sensitivity to quantization, the actual sensitivity also depends on the interplay of architectural design, feature representation, quantization methodology, and how these factors interact with each other.

\subsection{Qualitative Results}
% \vspace{-3mm}

Figure \ref{fig:Qualt} depicts  the impact of quantization on CAMs for the chosen deep learning models. Each row represents a different model (DenseNet-121, EfficientNet-B0, MobileNetV2, ResNet-50, SqueezeNet-1.0, VGG-16), and each column represents the effect of a different quantization level (f32, int16, int8). The first column shows the original image and the second  a ground truth segmentation of the salient object.

\textbf{VGG-16} demonstrates a gradual shift in the perceptual field,  where with the decrease of the quantization level, it expands its perceptual field, thus introducing more information noise about other parts of the image. 

\textbf{ResNEt-50} shows more robustness at the \textbf{int 16} quantization level, but a significant change in the perceptual field areas at the \textbf{int8} level. 

\textbf{DenseNet-121} shows similar behavior as \textbf{ResNEt-50}, but we notice that it tends to use a larger perceptual field covering more area of the stimuli.

\textbf{MobileNet-v2} shows less robustness  to  quantization effects, where the perceptual field changes at all precision levels. We also notice that the network tends to have a larger and more uniform perceptual field. It also tends to heavily emphasise on larger areas of the image as shown by the heatmap. 

In accordance with the empirical data, \textbf{SqueezeNet-1.0} is the least affected by the quantization, we can notice that it tends to have uniforme and more moderate coverage of the image maximizing the amount of innformation gathered, this explains it's hardiness against the loss of information stemming from quantization.

The \textbf{EfficientNet-B0} network shows the least robustness in accordance with the empirical data. The shift in the perceptual fields across different quantization levels is evident through the figure.

In summary, our comprehensive analysis reveals the profound impact of quantization on the perceptual fields of various neural network architectures, where both quantitative and qualitative analysis align their findings.

% By providing a detailed understanding of how quantization influences CAMs and model interpretability, this work contributes valuable insights for the development of more efficient and reliable neural network models. Future research can build on these results to optimize quantization techniques further, ensuring that model performance and interpretability are maintained across different deployment scenarios.

% Figure \ref{fig:Qualt} depicts an example of an image from the study dataset with the related overlaid salient object map. It also depicts the different CAMs associated with different models and levels of precision. 

% $VGG-16$, $ResNet50$, and $DenseNet121$ exhibit a low level of impact in terms of changes on CAMs, at the \textbf{int16} precision. On the other hand, we notice significant shifts when comparing at the \textbf{int8} quantization level. Which is consistent with the deductions from the empirical results. 

% Smaller networks like $MobileNet$, and $SqueezeNet$ show fewer changes across lower levels. We can especially notice that CAMs from $SqueezeNet$ remain unchanged. 

% $EfficientNet$ displays the highest level of changes across the models which means that it is much less robust. This is due to the simplicity of the network architecture and the small number of parameters. 

% \vspace{-3mm}

\section{Conclusion}
\label{sec:conclusion}
\vspace{-3mm}

This study investigated how quantization affects different neural network architectures using Class Activation Maps (CAMs) to gauge their alignment with human perception and the semantic content of the stimulus. 

The study demonstrates that while some models, such as SqueezeNet-1.0, exhibit remarkable robustness to quantization, others, like EfficientNet-B0, show significant sensitivity, particularly at lower precision levels. These findings underscore the importance of considering quantization effects when deploying deep learning models on resource-constrained devices especially in use cases of interpretability. 

These findings offer valuable insights for deploying neural networks on resource-constrained devices. By understanding how different architectures respond to quantization, researchers and practitioners can make informed decisions to achieve a balance between model performance, interpretability, and efficiency in real-world applications. Nonetheless, we need to mention that these results are still limited and that it would be beneficial to expand these results to more diverse datasets, models, domains, and tasks.

% This study delved into the impact of neural network compression on the perceptual capabilities of the network. Through the computation of Class Activation Maps (CAMs) to discern the pivotal regions influencing decisions, and juxtaposing them across various precision levels alongside visual saliency, we sought to discern the alterations within the network.

% Our investigation unveiled that neural networks engineered with a high number of parameters (e.g., VGG, ResNet, etc.) exhibit lower susceptibility to changes induced by compression. Conversely, networks crafted with a smaller parameter count demonstrate shifts in their spatial perceptual behavior at lower compression levels. This can open the hypothesis that over-parameterization can reduce the effect of compression on the perceptual ability of neural networks.   

% References should be produced using the bibtex program from sui table*
\small
\bibliographystyle{IEEEbib}
\bibliography{strings,refs}

\end{document}